\documentclass{article}

\usepackage[utf8]{inputenc}
\usepackage[T1]{fontenc} 
\usepackage[bookmarks=false]{hyperref}
\usepackage{url} 
\usepackage{booktabs}
\usepackage{amsfonts}
\usepackage{nicefrac}
\usepackage{microtype}
\usepackage{xcolor}
\usepackage{authblk}
\usepackage{amsmath}
\usepackage{amsfonts}
\usepackage{graphicx}
\usepackage{subfigure}
\usepackage{wrapfig} 
\usepackage{caption}
\usepackage{tabularx}
\usepackage[shortlabels]{enumitem}
\usepackage[margin=0.93in]{geometry}
\usepackage[parfill]{parskip}

\usepackage[style=numeric, sorting=none]{biblatex}
\title{Learning Dynamical Systems from Noisy Sensor Measurements using Multiple Shooting}

\author[1]{Armand Jordana}
\author[2]{Justin Carpentier}
\author[1]{Ludovic Righetti}
\affil[1]{Tandon School of Engineering, New York University, Brooklyn, NY}
\affil[2]{Inria, Département d’informatique de l’ENS, \'Ecole normale supérieure, CNRS, PSL Research University, Paris, France}

\date{}

\begin{document}

\maketitle

\begin{abstract}
Modeling dynamical systems plays a crucial role in capturing and understanding complex physical phenomena. When physical models are not sufficiently accurate or hardly describable by analytical formulas, one can use generic function approximators such as neural networks to capture the system dynamics directly from sensor measurements. As for now, current methods to learn the parameters of these neural networks are highly sensitive to the inherent instability of most dynamical systems of interest, which in turn prevents the study of very long sequences. In this work, we introduce a generic and scalable method based on multiple shooting to learn latent representations of indirectly observed dynamical systems. We achieve state-of-the-art performances on systems observed directly from raw images. Further, we demonstrate that our method is robust to noisy measurements and can handle complex dynamical systems, such as chaotic ones.

\end{abstract}
\section{Introduction}

\paragraph{Motivation.}
Learning dynamical systems offers the possibility to forecast the future, which plays a fundamental role in our understanding of complex physical phenomena in a large variety of fields: physics, chemistry, biology, engineering. An accurate model of the system dynamics also plays a central role in optimal control or model-based reinforcement learning~\cite{bertsekas2019reinforcement}. In real applications, model learning has to be done with measurements that are usually partial, indirect and corrupted by noise, and, consequently, a latent representation is needed. In this work, we introduce a generic approach to train neural networks forming a \textit{discrete nonlinear deterministic state space model} (SSM). Such representations have been widely studied in the system identification and control communities with parametric physical models and offer many beneficial properties for analysis and control synthesis. However, due to the resulting complex maximum likelihood estimator, it remains challenging to learn such representations with deep neural networks. 
This paper shows that state space representations are compatible with neural networks and proposes a generic, yet computationally simple way to learn them. Our framework can consider complex measurements such as raw images and can capture complex dynamics such as chaotic dynamics.

\paragraph{Problem statement.}
A \textit{discrete nonlinear deterministic SSM} can be written in the following way:
\begin{subequations}
\begin{align}
    x_{t+1} &= f_{\theta}(x_t)\\
    y_t &= g_{\theta}(x_t) + \epsilon_t
\end{align}
\end{subequations}
where $x_t$ is the state vector at time $t$ and $y_t$ the measurement vector, also referred as the observation. The measurement noise is modeled by the random variables, $(\epsilon_t)_t$, which are mutually independent conditionally to the initial condition and model parameters.
Hence, the state is Markov -- only the current state is necessary to make predictions in the future. The transition function,  $f_{\theta}$, and the measurement function, $g_{\theta}$, are both represented by neural networks parametrized by the vector~$\theta$. One can interpret $g_{\theta}$ as a decoder and $x_t$ as a latent state. Our goal is to infer $\theta$ given different sequences of measurements $ (y_1, ..., y_T)$. The main challenge is that $g_{\theta}$ is usually not bijective and a single measurement $y_t$ does not provide sufficient information to infer the current state $x_t$. Instead, a sequence of measurements is necessary to infer the state and the model parameters~$\theta$. Interestingly, the problem still occurs when the measurement function $g_{\theta}$ is equal to the identity mapping. If one tries to train the transition network in a supervised way with pairs of consecutive measurements, both the input and the targeted output of the network will be corrupted by noise and this will introduce bias in the estimation. This is the so-called Error-in-Variable problem  which has been extensively studied in the control community~\cite{jaeger1996unbiased, voss2004nonlinear}. In the model-based reinforcement learning literature~\cite{levine2014learning, chua2018deep, bechtle2020curious}, this problem is often ignored and data just filtered to reduce  noise. However, even with filtered data, direct supervised training of the transition model remains biased. Our approach follows the maximum likelihood estimator such that this bias can be avoided despite the use of neural networks. 
\paragraph{Related work.}
There exists a vast literature on the learning of dynamical systems. We refer to \cite{brunton2019data} for a detailed overview. In this work, we are interested by the partial observability setting and will focus this review on work addressing this scenario. Additionally, we review relevant work on video prediction and learning of chaotic systems such as the Lorenz system.

In the system identification community, the problem of computing an unbiased SSM from data is often cast as an Initial Value Problem (IVP) and can be solved using multiple shooting optimization techniques. These methods, widely studied,  allow a rigorous inference of the system parameters~\cite{Bock1983, timmer1998modeling, voss2004nonlinear, peifer2007parameter}. While \cite{ribeiro2020smoothness} has shown that small neural networks can be trained with multiple shooting, their approach cannot scale to large networks and complex measurements such as images. In particular, their problem formulation prevents the use of stochastic gradient descent techniques. In this paper, we show how IVP and multiple shooting can be advantageously exploited to learn complex models with large networks.

Recent works have focused on recovering the latent dynamics of partially observed systems through the scope of variational autoencoders~(VAE) and recurrent neural networks (RNN). These networks are trained by maximizing the evidence lower bound~\cite{krishnan2015deep, karl2016deep, fraccaro2017disentangled, krishnan2017structured, naesseth2017variational, duncker2019learning}. Those works assume that the dynamics are stochastic. In contrast, we are interested by systems with deterministic dynamics. Thanks to our multiple shooting formulation, the inference can be done without approximating the maximum likelihood estimator through the lower bound and without the use of a recurrent encoding network. While the stochastic assumption can be relevant in fields such as biology or statistical physics, our main interest is to learn mechanical systems, such as robots, that can accurately be modeled with deterministic dynamics and noisy sensors. Hence, we believe it is crucial to leverage this fact to our advantage, and to also later enable the use of well established analysis and control synthesis techniques with the learned SSM. %

Another line of work is to replace the latent dynamics learning with a RNN mapping sequences to sequences such as LSTM \cite{hochreiter1997long} or GRU \cite{cho2014properties} and train in a supervised way. \cite{becker2019recurrent} introduced Recurrent Kalman networks (RKN), a RNN with gates handling the measurement noise as a Kalman filter. A large state is necessary in order to match the linearity assumptions. Therefore, the latent state cannot be interpreted, as it is not possible to directly match some components of the latent state to the observations. In contrast, our approach enables learning low dimensional and compact  state representations where some components may be forced to have some physical meaning. In addition, RKNs do not learn a state space representation but an inference framework. In our experiment, we will compare our method to both RKN and LSTM. However, our approach is conceptually very different. While LSTM and GRU can handle long term dependencies, this ability relies on a complex design of units to avoid the vanishing gradient problem. In contrast, our method allows any sort of architecture because long term dependencies are handled by the multiple shooting strategy.

There has been a significant interest in the learning of the dynamics of the Lorenz system where the state is observed with or without noise \cite{guerra2008multi, quade2018sparse, rudy2019deep}. Regarding the partially observable setting, \cite{hernandez2018nonlinear, zhao2020variational} reconstruct the Lorenz attractor given high dimensional observation with a linear observation model. \cite{brunton2017chaos} study in depth the Lorenz system given a one dimensional measurement with time delayed embedding. In contrast to those works, we learn a 3 dimensional SSM given a scalar measurement. Our work is most similar to \cite{mirowski2009dynamic} which also learns a discrete nonlinear deterministic SSM although they restrict their study to linear observation models only. Hence, our multiple shooting formulation can be seen as an extension of their work.

Other methods have been developed in order to tackle more specifically the video prediction problem~\cite{finn2016unsupervised, babaeizadeh2017stochastic, kalchbrenner2017video, hsieh2018learning, guen2020disentangling}. \cite{srivastava2015unsupervised} proposed a LSTM encoder-decoder framework to learn video representations.  \cite{minderer2019unsupervised} builds a keypoint-based image representation and learns a stochastic dynamics model of the key-points. Finally, \cite{mathieu2015deep, tulyakov2018mocogan, lee2018stochastic} used generative adversarial networks to perform video prediction. In contrast to those methods, our SSM is not restricted to image measurements.

\vspace{-1mm}
\paragraph{Contributions.}
We introduce an algorithm to directly learn {\em discrete nonlinear deterministic state space models}. We cast the training of the transition and observation networks as an IVP: given a sequence of measurements, we find the optimal unbiased initial conditions and network parameters that can reproduce the sequence. Importantly, we present a scalable and generic multiple shooting formulation that stabilizes training of the IVP over long trajectories while affording parallelization. We show that our method can learn the latent dynamics of systems observed directly from raw images and achieve a prediction accuracy similar to the state of the art. To the best of our knowledge, this is the first demonstration that multiple shooting can be used for the estimation of dynamics given raw images. In addition, we demonstrate that our approach can handle very long temporal series. We show the ability of our method to accurately learn a partially observed chaotic system where up to ten thousand time steps are required for precise inference.
Conceptually, our method provides a link between state space representations from the control and identification literature and the notion of latent representations common in machine learning. It can handle low dimensional state representation with highly nonlinear dynamics or high dimensional states with locally linear dynamics. A compact representation may be useful to retain a level of interpretability of the latent dynamics, while locally linear dynamics can simplify feedback control design.

\section{Background}
\label{section:background}
\paragraph{Initial Value Problem.}
We now present the IVP and the loss to infer the parameters of the SSM. Given a sequence of measurements $(y_1, ..., y_T)$, the maximum likelihood estimate can be found by maximizing $p(y_1, ..., y_T | \theta)$. In general, no prior on $p(x_1)$ is available, therefore, it is assumed that $p(x_1)$ follows a uniform prior distribution. Thus, the estimation is done by maximizing $p(y_1, ..., y_T | \theta, x_1)$ jointly on $\theta$ and $x_1$. As the noise is assumed to be mutually independent conditionally to the initial state and parameters, the following factorization holds:
\begin{align}
p(y_1, ..., y_T |\theta, x_1) &= \prod_{t=1}^{T}  p(y_t|\theta, x_1) \label{indep}
\end{align}
Because the dynamics is deterministic, the measurement $y_t$ only depends on $x_t$ which is nothing but the $t-1$ iteration of the transition network: $f^{(t-1)}_{\theta}(x_1)$. Each random variable, $\epsilon_t$, is assumed to follow a centered Gaussian distribution with a fixed diagonal covariance of the form $\sigma I$. By applying the negative logarithm, one can show that the problem is equivalent to the following minimization:%
\begin{align}
    \displaystyle \underset{x_1, \theta}{\operatorname{argmin}} \sum_{t=1}^{T}\| g_{\theta}(f^{(t-1)}_{\theta}(x_1)) - y_{t}\|_2^2.
    \label{eq:ivp}
\end{align}
This minimization problem is called the IVP. It is important to notice that we optimize both the network parameters $\theta$ and the initial state $x_1$, necessary to recover an unbiased model. As $g_{\theta}(f^{(t-1)}_{\theta}(x_1)) $ is equal to $g_{\theta}(x_t)$, the intuition of the IVP is that we look for a latent trajectory that best generates the targeted measurements. Thanks to the deterministic dynamics assumption, the search can be done only on the initial condition without requiring an encoding network.
One of the difficulty when solving IVPs lies in the landscape of the cost function which presents several local minimums due to the successive iterations over the transition model~\cite{voss2004nonlinear}. Even for a linear transition model the cost function is non-convex. Furthermore, iterating too many times an untrained network will necessarily blow up at some point. For these reasons, the IVP is limited to trajectories spanning a short time horizon. Yet, many dynamical systems require long time horizons to be inferred precisely.

\paragraph{Multiple Shooting.}
The usual way to circumvent the limitations of the IVP is to lift the optimization problem into a higher dimensional space by introducing slack variables and equality constraints in order to break down the complexity of the successive iterations. This is the purpose of multiple shooting~\cite{Bock1983}. The core idea is to split the trajectory in $m$ sub-trajectories of length \mbox{$n = \frac{T}{m}$} and search for the initial conditions of each sub-trajectory with equality constraints guaranteeing the continuity of the trajectory. This can be formalized in the following way:
\begin{subequations}
\begin{align}
  \underset{s_1, ...,s_{m} , \theta}{\operatorname{argmin}} & \sum_{i=1}^{m} \sum_{k=1}^{n}\| g_{\theta}(f^{(k-1)}_{\theta}(s_{i})) - y_{(i-1) n + k}  \|_2^2    \\ 
  \mbox{  s.t.  } s_{i+1} &= f_{\theta}^{(n)} (s_{i}) \;\;\;\;\; 1 \leq i \leq m-1
  \label{ms_hard}
\end{align}
\end{subequations}
The first sum is done over the different sub-trajectories while the second sum is done over the time steps of each sub-trajectory. The initial conditions of each sub-trajectories, $s_i$, are called the shooting nodes. This optimization problem is mathematically equivalent to the original IVP~\eqref{eq:ivp}. The constraints guarantee that, put together, each sub-trajectory forms one long trajectory where for each time step: \mbox{$x_t = f^{(t-1)}_{\theta}(x_1)$} when the optimization process has converged on the constraints. In addition, if only one sub-trajectory is considered, $m=1$, we recover the original IVP. The parameter $n$ allows to choose the maximum number of iterations of the transition model. A high value for $n$, above a hundred for instance, might create some numerical instability during training. Indeed, trajectories can diverge exponentially through successive iterations of the transition model~\cite{venkatraman2015improving, ribeiro2020smoothness}. However, a too low value might not be numerically efficient as the whole problem would be learned through the constraints. In practice, we choose $10 \leq n \leq 100$.
\begin{figure}[ht]
\begin{center}
\centerline{\includegraphics[trim=10 10 10 10, clip, width=0.9\textwidth]{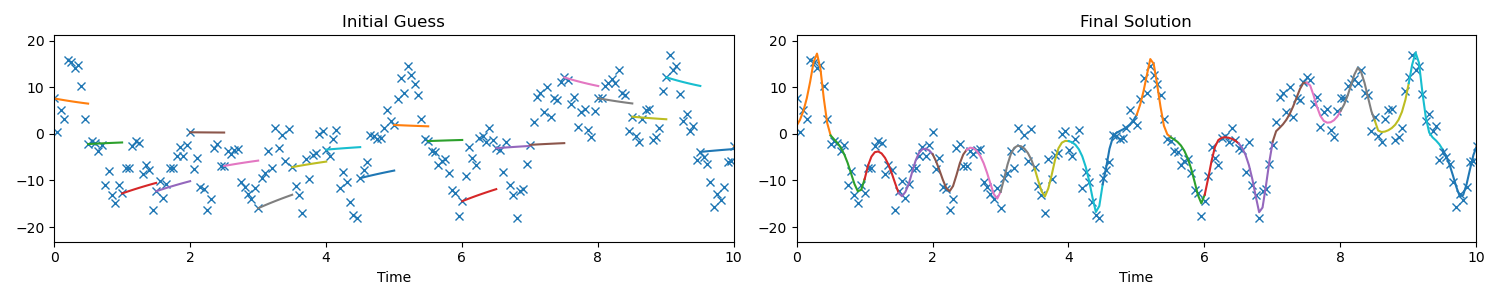}}
\caption{Schematic of the multiple shooting approach in the simplified case where $g_{\theta}$ is the identity mapping. Each dot corresponds to a measurement while each curve corresponds to a sub-trajectory.} 
\label{schematic}
\end{center}
\end{figure}

Figure~\ref{schematic} presents a schematic overview of the multiple shooting approach in the simplified case where $g_{\theta}$ is the identity mapping.
We plot a sequence of measurements of the Lorenz system and the corresponding sub-trajectories. 
There are twenty sub-trajectories of ten time steps each. The shooting nodes are initialized at the measurement of the corresponding time step. At the beginning, the transition network is not trained yet and the sub-trajectories do not follow the dynamics. The gaps between each sub-trajectory correspond to the value of the continuation constraints. %
After training, a trajectory without noise is recovered. Here, training is done with several trajectories. It is important to notice that with only one trajectory, the transition model would likely overfit to the noise.
One of the key benefits of the multiple-shooting method is that it allows to provide extra knowledge at the initialization of the shooting nodes: a guess on the initial condition of each sub-trajectory can be taken. Information is provided throughout the whole trajectory while the IVP can provide information only at the initial condition. In practice, this stabilizes and speeds up the training. When no knowledge on the value of the shooting nodes is available, they can be initialized to zero. 

\section{Method}
\label{section:method}

\paragraph{Model inference.}
We now show how multiple shooting can be used with neural networks. There are two challenges: the equality constraints and the minimization over the shooting nodes. To enforce the constraints, we transform the constrained optimization problem\,\eqref{ms_hard} into an unconstrained problem thanks to a L2 penalty term~\cite{nocedal2006penalty}. For each trajectory, the loss becomes:
\begin{align}
  l(s_1, ...,s_{m} , \theta) =& \frac{1}{mn}\sum_{i=1}^{m} \sum_{k=1}^{n}\| g_{\theta}(f^{(k-1)}_{\theta}(s_i)) - y_{(i-1) n + k}  \|_2^2 
  +  \frac{\alpha}{m-1} \sum_{i=1}^{m-1} \| s_{i+1} - f_{\theta}^{(n)} (s_i) \|^2_2
\label{penalty}
\end{align}

The first term corresponds to the original cost while the second term enforces the soft constraint where $\alpha$ is the penalty parameter. More complex optimization procedures such as an augmented Lagrangian %
could be considered but the penalty method already yields satisfying results. Generalization to $N$ trajectories is done by introducing a set of shooting nodes for each trajectory and summing the loss:
\begin{align}
    \mathcal{L}((s_i^j)_{i,j}, \theta) &= \frac{1}{N} \sum_{j=1}^{N} l(s_1^j, ...,s_{m}^j , \theta)
\end{align}
where $s_i^j$ corresponds to the $i^{th}$ shooting node of the $j^{th}$ trajectory. This loss is minimized with stochastic gradient descent. The optimization is done jointly on the weights of the networks and the shooting nodes. More specifically, each shooting node is considered as a trainable variable in the same way as the weights. Each trajectory is considered as one data sample, therefore, each shooting node is updated when its corresponding trajectory is selected in the batch. This happens one time per epoch. In practice, the training behaves well as long as the stochasticity of the gradient descent is preserved -- the number of trajectory per batch is much smaller than the total number of trajectories.

Theoretically, the penalty parameter should be increased indefinitely throughout training~\cite{nocedal2006penalty}. In practice, we initialize $\alpha$ with the value $1$ and increase it once, after 200 epochs. At the same time, we reinitialize the optimizer and reduce the learning rate by 10. Lastly, we impose an other learning rate decay after 600 epochs. As each shooting node is updated once per epoch, the total number of updates of the shooting nodes is equal to the total number of epochs. Consequently, this training procedure requires a minimum number of epochs, we find that in practice, one thousand is enough for the more complex scenarios.
In addition to numerical stability, the multiple-shooting approach brings computational efficiency. Indeed, the loss and penalty on each sub-trajectory can be computed independently. Therefore, the computations can be performed in parallel on a GPU. %

\paragraph{Latent state inference.}
Our problem formulation shows that the SSM can be learned without a recurrent encoder, i.e. a mapping from the measurement space to the state space. This brings two benefits: it simplifies the optimization procedure and it offers the possibility to use any latent state inference tool. In practice, a state inference tool could be used to forecast the future evolution of unseen trajectories. More specifically, given $(y_1, ..., y_t)$, one may want to predict $(y_{t+1}, y_{t+2}, ...)$. To do so, it is necessary to find a state $x_t$ given the previous measurement $(y_1, ..., y_t)$ in order to predict the next measurements by iterating the transition network and applying the observation network to the next predicted states. This state inference problem is easily solved with a probabilistic filter. We refer to~\cite{thrun2002probabilistic} for a detailed overview of the topic. For instance, an unscented Kalman filter (UKF) \cite{wan2000unscented} or a particle filter can be used to recover the latent state given previous observations. These methods only rely on the transition and observation networks to do the inference. Another possibility is to train a recurrent encoder mapping $(y_1, ..., y_t)$ to $x_t$. More specifically, after the training of the SSM, the shooting nodes and the transition model can recover each state trajectory and an RNN mapping  $(y_1, ..., y_T)$ to  $(x_1, ..., x_T)$ can be trained in a supervised manner. The strength of the SMM formalism is that any inference tool can be used. In our experiments, for measurements other than images, we use UKFs. For images, a particle filter can be used, however, we find that using a CNN-LSTM yields better performances. This is not surprising as CNN are very good at extracting the relevant information of an image. Conceptually, training a RNN is not ideal as the recurrent units are to some extent learning again the dynamics of the system. A more elegant solution would be to use a CNN to extract the relevant features and then a probabilistic filter similar to \cite{becker2019recurrent}. However, this is beyond the scope of the paper.%

\section{Experiments}
\label{section:experiments}

In this section, we first show that our method learns accurate latent dynamics from images. Then, we demonstrate the robustness of our approach when dealing with chaotic dynamics such as the Lorenz  system. Further, we use this system to highlight the fact that standard probabilistic filters can be used for state inference. A Python implementation of the multiple shooting framework is available online\footnote{\url{https://github.com/machines-in-motion/ssm_multiple_shooting}}. We use Pytorch for the implementation \cite{paszke2019pytorch} and Adam for the optimization. In order to demonstrate the robustness and versatility of our method, we study two different types of transition model: locally linear models \cite{karl2016deep, fraccaro2017disentangled} and fully connected models. Locally linear models (SSM-LL) are of the form:
\begin{align}
    f_{\theta}(x) = x + \sum_{k=1}^K \beta_k(x) A_k x 
\end{align}
where $\beta_k$ is the $k^{th}$ output of a neural network outputting a vector of size $K$ with one hidden layer and a softmax activation after the last layer. Here, $\theta$ contains the weights of $\beta$ and each $A_k$. In contrast, fully connected models (SSM-FC) are of the form:
\begin{align}
    f_{\theta}(x) = x + \gamma(x)
\end{align}
where $\gamma$ is a fully connected network. Note that for both models, we add the term $x$ in the model equation. Consequently, the networks learn to output $x_{t+1} - x_t$ instead of $x_{t+1}$. Indeed, for most dynamical system the state trajectory evolves continuously, therefore, $\|x_{t+1} - x_t\|$ is usually small. As approximating the zero function is easier than approximating the identity map, this helps making the model more expressive. Lastly, in each of our experiments, the weights of the transition and observation networks are distinct and $\theta$ is the collection of parameters of both networks. 

\paragraph{Simple pendulum.}
In the first experiment we consider the simple pendulum observed through $24 \times 24$ images. We generate one thousand training trajectories of 10\,s with a time step of 0.1\,s, yielding trajectories of 100 time steps each. Each image is corrupted with Gaussian noise with a standard deviation of 0.2 given pixels between 0 and 1. To generate an accurate image, the decoder $g_{\theta}$ is made of one fully connected layer and three transposed convolutions. Further details on the networks architectures and the dataset can be found in the supplementary material. In this experiment, the measurement is totally indirect and no prior information on the value of shooting nodes is available. Therefore, each shooting node is initialized to zero. We train for one thousand epochs with 4 shooting nodes per trajectories. After the training of the transition and observation models, we use a RNN to solve the state inference task: we train a CNN-LSTM that maps each training sequence of images to the learned latent trajectory. Then, on each testing sequence, given the first 50 images, the RNN provides the latent state at time step 50 which we use to predict the next 50 images.

In Table \ref{simple-table}, we report the average mean squared error (MSE) and binary cross entropy (BCE) per pixel on one thousand testing trajectories. The average is done on both the total number of pixels and the total number of predicted frames. We compare to RKN and to a sequence-to-sequence CNN-LSTM (referred as LSTM in  Table~\ref{simple-table})  which report state-of-the-art results for such tasks and we observe similar performances. 
Both those methods are trained with the same CNN architecture as the one used in our filtering RNN. By nature, those methods need an input at each time step, even during prediction. This is counter intuitive as during prediction, there are no measurement available. To circumvent this issue, we follow \cite{becker2019recurrent} and use boolean values to distinguish between filtering and prediction phases and the measurement are artificially set to zero during the prediction phase. It is important to note that the RNN that we use for filtering in our method does not need this additional boolean input as the RNN is not used for prediction. Forecasting is entirely done by iterating the SSM.

\begin{table}[ht]
\caption{Prediction errors for the simple pendulum}
\label{simple-table}
\begin{center}
\begin{tabular}{|c|c|c|}
\hline
& MSE & BCE \\
\hline
SSM-FC &  $0.0154  \pm   0.0026$  &  $0.2902 \pm 0.0075 $ \\
\hline
SSM-LL &  $  0.0152   \pm   0.0022 $  &  $0.2895  \pm   0.0064$ \\
\hline
LSTM &  $ 0.0158 \pm 0.0034 $  &  $ 0.2913 \pm 0.0098 $\\
\hline
RKN &  $ 0.0155 \pm 0.0021 $  &  $ 0.2902  \pm  0.0058 $\\
\hline
\end{tabular}
\end{center}
\end{table}

\begin{figure}[ht]
\centerline{\includegraphics[width=0.8\columnwidth]{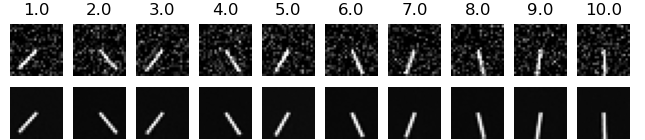}}
\caption{Prediction on the simple pendulum. The first row corresponds to the noisy input and the second to the prediction. The prediction starts at time 5 after the filtering process.}
\label{pendulum_test}
\end{figure}

In this experiment, we find that a latent space of dimension 10 performs well. However, smaller latent spaces can also be considered. As consequence, the resulting model would be more explainable and mode suitable for control, while having only a slight decrease in performances. For instance for a 3D latent state and a locally linear model, we find a MSE of  $0.0177   \pm  0.0043$ and a BCE of $0.2968   \pm   0.0120$. Quantitatively and qualitatively, the results are hardly distinguishable. In Figure~\ref{pendulum_test}, a testing sequence with one frame out of ten is plotted. Before time 5, the RNN is provided the ground truth images and infers a latent state trajectory that we map to a sequence of images with the observation network. Then, we use the latent state at time 5 to predict the next 5\,s. We observe that the predicted sequence does not diverge from the ground truth, demonstrating the accuracy the transition network after 50 iterations. Additionally, the sharpness of the images shows the ability of the observation network to learn a mapping to the measurement space while removing noise.

\paragraph{Quadruple pendulum.}

\begin{figure}[ht]
\centerline{\includegraphics[width=0.78\columnwidth]{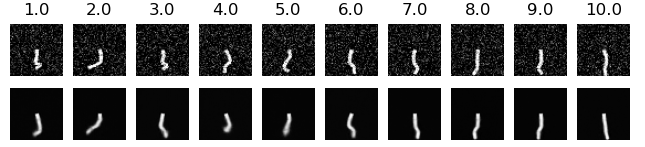}}
\captionof{figure}{Prediction on the quadruple pendulum. The first row corresponds to the ground truth and the second to the prediction. The prediction starts at time 5 after the filtering process.}
\label{4pendulum_test}
\end{figure}

To show the robustness of our method to complex dynamics, we benchmark our method on the quadruple pendulum. The dynamics are simulated with Pinocchio~\cite{pinocchioweb} and we generate 2000 training trajectories of 100 times steps with $48 \times 48$ images that we corrupt with Gaussian noise. The observation network is made of one fully connected layer and four transposed convolutions and we consider a latent state of dimension 40. We train for one thousand epochs with 4 shooting nodes per trajectories. In the same way as for the previous experiment, a RNN is used to infer the latent state during evaluation. Given the first 50 time steps, we predict the next 50 images on 2000 testing trajectories. In Table \ref{multiple-table}, we report the average BCE and MSE. Again, we perform similarly to RKN and LSTM. Figure \ref{4pendulum_test} shows a testing sequence (one frame every ten). In the same way as for the simple pendulum, the first 5 seconds correspond to state inference while, the last 5 seconds correspond to prediction. The prediction is accurate for several seconds which is non trivial for a system of such complexity.
\begin{table}[ht]
\caption{Prediction errors for the quadruple pendulum}
\label{multiple-table}
\begin{center}
\begin{tabular}{|c|c|c|}
\hline
& MSE & BCE \\
\hline
SSM-FC & $ 0.0250 \pm 0.0053$  & $0.3119 \pm  0.0140$ \\
\hline
SSM-LL &   $0.0245   \pm  0.0055$  & $ 0.3109  \pm   0.0146 $ \\
\hline
LSTM &   $ 0.0252 \pm 0.0073 $  & $ 0.3161 \pm 0.0219 $ \\
\hline
RKN &  $ 0.0243  \pm  0.0059$  &  $  0.3108  \pm 0.0160$\\
\hline
\end{tabular}
\end{center}
\end{table} 

\paragraph{Moving MNIST.}
We now evaluate our method on the Moving MNIST dataset \cite{srivastava2015unsupervised} to demonstrate that our approach can cope
with complex measurement models and large datasets.
With this dataset, training sequences can be generated on-the-fly by sampling MNIST digits which enables the use of a virtually infinite dataset for training. For example, \cite{kalchbrenner2017video} uses 19.2 millions training trajectories.
This is, however, not necessarily realistic for real-world applications.
By design, our method uses a finite training set and we consider here 100,000 training trajectories, each made of 20 images of dimension $64\,\times\,64$. We train a SSM with a 100 dimensional latent state, a locally linear transition model and two shooting nodes per trajectory. Once the SSM is learned, we train a CNN-LSTM for state inference. 

\begin{figure}[ht]
\centerline{\includegraphics[width=0.79\columnwidth]{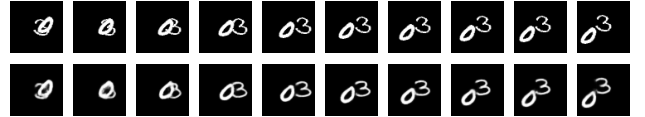}}
\caption{10 predicted frames on Moving MNIST (first row: ground truth - second row: prediction).}
\label{MovingMNIST_test}
\end{figure}

Figure~\ref{MovingMNIST_test} shows a typical example of 10 predicted frames. While the structure of the dynamics is clearly understood, we notice that the shape of the digits are slightly different from the ground truth (i.e. the measurement model could be improved). 
This artifact can be explained by the limited size of the training dataset.
Our current implementation and cluster access limits our dataset to 100,000 trajectories as we currently load all the dataset in memory and have limited memory resources for training.
Nonetheless, as reported in Table~\ref{MNIST-table}, our approach still performs reasonably well with performance comparable to \cite{srivastava2015unsupervised}. While we do not reach the state of the art achieved by \cite{kalchbrenner2017video}, they need 2 orders of magnitude more training data to achieve this performance.
It is also worth noting that all methods performing better are specifically tailored to the video prediction problem while our approach does not make any assumption on the nature of the measurements and system dynamics.
This experiment demonstrates that our method easily scales as we used a total of 2,000,000 images (of size 64 by 64) and 200,000 shooting nodes without any numerical issues. To the best of our knowledge, no other methods based on IVP and multiple-shooting are capable of handling this number of trajectories and shooting nodes. %

\begin{table}[ht]
\caption{Prediction errors for Moving MNIST For this experiment only, we follow \cite{srivastava2015unsupervised} and do not average the metric other the total number of pixels }
\label{MNIST-table}
\begin{center}
\begin{tabularx}{0.3\textwidth}{ 
  | >{\centering\arraybackslash}X 
  | >{\centering\arraybackslash}X | }
\hline
& BCE \\
\hline
\cite{shi2015convolutional}& $367.2 $\\
\hline
Ours & $353.3$\\
\hline
\cite{srivastava2015unsupervised} &  $341.2$ \\
\hline
 \cite{kalchbrenner2017video} & $87.6$ \\
\hline
\end{tabularx}
\end{center}
\end{table} 

\paragraph{Lorenz system.}
In the last experiment, we show how our approach can learn intricate dynamics requiring long temporal sequences under partial noisy observations. We consider the Lorenz system:
\begin{equation}
    \dot x = \sigma (y - x) \;\;\;\;\;\;\;\;\;
    \dot y = x (\rho - z) - y \;\;\;\;\;\;\;\;\;
    \dot z = x y - \beta z  
\end{equation}
where $ (\sigma, \beta, \rho) = (10, 8/3, 28)$ ensure the dynamics is chaotic.
We only use as measurement the first component of the state, $x$, that we corrupt with a Gaussian noise with a standard deviation equal to 30\% of the standard deviation of~$x$. It is notoriously difficult to distinguish chaos (a deterministic behavior) from noise \cite{rosso2007distinguishing}.
In a chaotic system, infinitely close trajectories diverge exponentially fast from each other so we do not expect to be able to perform accurate long term predictions. However, the system possesses a stable structure, its strange attractor \cite{strogatz2018nonlinear}, which should be recoverable. The training set is made of 1000 trajectories of 50s with a 5ms time step. %
We chose a latent state of three dimensions and exploit the knowledge that the measurement is a component of the state by imposing the observation function to be the projection mapping, 
$g_{\theta}(x_1, x_2, x_3) = x_1$. Consequently, this forces the first component of our state representation to be the same as the x-component of the real system.
This illustrates how prior knowledge about the nature of the observations can be easily included in our framework.
We initialize the first coordinate of each shooting node at the value of the corresponding measurement of the $x$-coordinate, while the two others components are initialized at zero. 

We use both a RNN and a UKF for state inference to demonstrate that our representation also affords the use of standard estimation techniques. To evaluate our model, we look at the prediction performance on 1000 testing trajectories of 2000 time steps. We train the RNN as in the previous examples. In contrast, the UKF directly uses the transition and observation networks to perform state inference. At each time step, the UKF scans each measurement and updates its prior on the state \cite{thrun2002probabilistic}. The filter is initialized as follows: the noisy measurement on the first coordinate and zero on the two other components as we have no prior for those coordinates. Then, at each time step, the transition and the observation networks are used to check if the predicted measurement matches the real next measurement. Given the error, an update on the prior is computed. For both inference frameworks, we use the first 1990 measurements of each testing sequence to infer the state. Then, we use the state with the transition and observation networks to predict the next 10 measurements. Only 10 steps prediction are used because of the chaotic nature of the system. Larger prediction might diverge and give a high MSE that could not be interpreted. The testing metric is the average MSE between the predicted sequence and the ground truth without noise.

\begin{figure}[ht]
\centering
\begin{minipage}{.5\textwidth}
\centerline{\includegraphics[trim=10 5 10 10, clip, width=0.9\columnwidth]{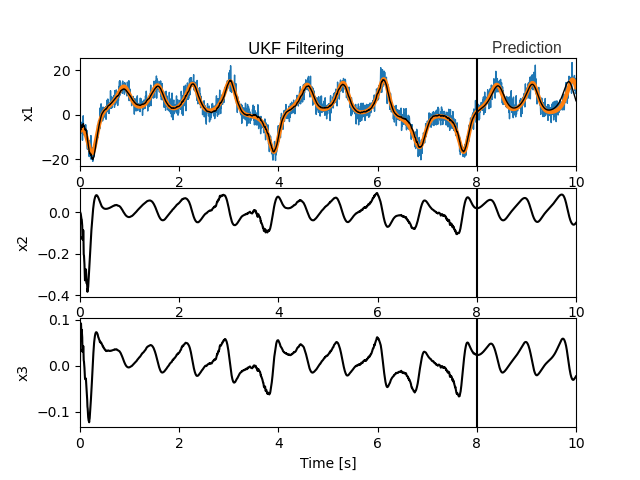}}
\captionof{figure}{Prediction on the Lorenz system. Each axis corresponds to one component of the state (black curve) where the first one is also the measurement. We also show the measurement (blue curve) and the ground truth (yellow curve). }
\label{lorenz_test}
\end{minipage}%
\hspace{0.3cm}
\begin{minipage}{.46\textwidth}
\centerline{\includegraphics[width=0.9\columnwidth, trim={0 0 0 1cm}, clip]{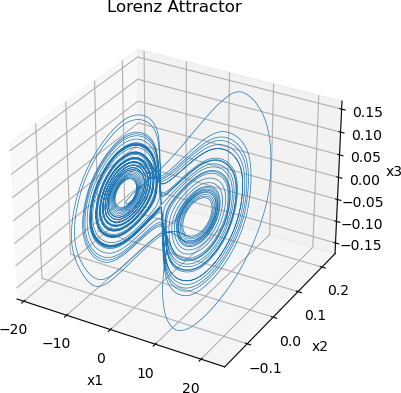}}
\captionof{figure}{Lorenz attractor. Each axis corresponds to one component of the state. After training,  we take a shooting node and iterate ten thousand times the transition model.}
\label{butterfly}
\end{minipage}
\end{figure}

Figure~\ref{lorenz_test} shows a typical result for a 2s prediction. The first 8 seconds are inference steps, and the last 2 seconds are prediction steps given the inferred state at time 8. There are three axis, one for each component of the state where the first one also corresponds to the measurement. Before time 8, the plotted state corresponds to the belief of the filter. This prediction is made by iterating 400 times the transition model and we can see that the corresponding measurement matches the ground truth.
Longer predictions are not meaningful as trajectories will diverge exponentially fast from the ground truth for any infinitesimal error. Typically, the prediction works well until a change of orbit occurs. However, our SSM is able to correctly learn the strange attractor (Figure~\ref{butterfly}), the stable structure of the Lorenz system. Thanks to the chosen observation function, the first component is indeed the same as the x component while the two others have been constructed by the learning process.
This result demonstrates the model's ability to capture the essential characteristics of complex physical phenomena and to learn qualitatively correct long term behavior. 

Finally, we demonstrate that because the system is partially observed and corrupted by noise, long sequences are indeed needed to accurately learn the SSM. We consider four different types of training datasets with trajectories of various lengths: 10, 100, 1000 and 10,000 time steps. In order to provide a fair comparison, we make sure that each dataset contains the same number of total samples, resulting in different numbers of trajectories between datasets. For the two datasets containing trajectories of length 1000 and 10000, we consider sub-trajectories made of 50 time steps, while for trajectories of length 10 and 100, we use sub-trajectories of length 10. For this experiment, we find that the fully connected network performs poorly and therefore, we only present the results of the locally linear model. In Table~\ref{lorenz-table}, we report the average prediction error for each dataset and each state inference framework. The lowest errors are obtained with the dataset containing the longest sequences. For state inference, we note that the RNN yields superior performances, however, the UKF works reasonably well and demonstrates that the transition and observation networks can directly be used for state inference. 
As the state is partially observed and corrupted by a high amount of noise, very long sequences are required to accurately learn a model as suggested by our results. This experiment makes the inference especially challenging for numerical stability. Small errors in the model create trajectories diverging very quickly from the ground truth and the use of multiple shooting is particularly important to handle long trajectories.
Note that we were not able to train RKN. As for the sequence-to-sequence LSTM, we do not achieve as good performances as our SSM. We believe that this is due to the fact that there is no way to take advantage of the structure of the problem -- the measurement is a noisy observation of a component of the state. Consequently, this structure needs to be learned by the LSTM which is not necessary and very complex because of the high variance of the noise.
\begin{table}[ht]
\caption{Prediction error for the Lorenz system for different length of training sequence.}
\label{lorenz-table}
\begin{center}
\begin{small}
\begin{tabular}{|c|c|c|c|c|}
\hline
T &  10 & 100  &  1000 & 10,000 \\
\hline
SSM-RNN &  $ 3592.273 $ & $ 10.284  $  &  $ 0.391 $ & $ 0.154 $\\
\hline
SSM-UKF &  $ 46.702 $ & $ 10.999 $  &  $  0.670 $ & $ 0.335 $\\
\hline
LSTM & $35.630$ &$34.852$& $6.867$ & $0.314$\\
\hline
RKN &  did not converge & did not converge & $67.76$ & out of memory\\
\hline
\end{tabular}
\end{small}
\end{center}
\end{table}

\section{Discussion and conclusion}

In this work, Gaussian noise is assumed with a covariance of the form $\sigma I$. This assumption might not hold when different types of sensors are considered. In that case, one can introduce a weighted loss for each different sensor \cite{voss2004nonlinear}. While we only considered the MSE loss as we worked with Gaussian noise, the multiple shooting approach is not restricted to such loss. If a different type of measurement noise was considered, the loss could be changed accordingly. For instance, the binary cross entropy loss could be used with images. Regarding the optimization procedure, the L2 penalty over the shooting nodes is fairly naive and requires tuning the penalty parameter. We anticipate that methods such as an augmented Lagrangian could avoid manually tuning the penalty parameter.

To conclude, we have shown how the inference of neural networks forming a SSM could be cast as an IVP and solved efficiently using multiple shooting. The method is generic and scales to large networks and long time dependencies. Additionally, our method allows a flexible architecture design. For the transition model, we notably showed that both locally linear models and fully connected models could be used. For the observation model, we used neural networks with transposed convolution to generate state-of-the-art image sequence predictions but also projection map to include measurements into the latent space. Further, this work suggests that learning latent dynamics can be done without RNN. The SSM can directly be trained with shooting nodes instead of a RNN and probabilistic filters can directly use the transition and observation networks to perform state inference. In future work, we will consider more sophisticated controlled systems such as robots using multi-sensor measurements.

\paragraph{Broader impact.}
This paper focuses on a general-purpose methodology to learn dynamical systems. We believe that the general goal of this paper does not raise ethical concerns as the learning of dynamical system aims to better understand physical phenomena. However, the applications that can result from this methodology are diverse and some can raise ethical concerns. For instance, the learning of dynamics could be used into weapons systems. Further, the proposed approach lacks formal guarantees that it will work in every conditions. We therefore do not recommend its use in safety-critical applications without
further performance analysis (e.g. in an autonomous driving system). Lastly, every dataset is synthetic and this paper does not consider any human-derived data.

\paragraph{Acknowledgment.} 
This work was supported in part by the French government under management of Agence Nationale de la Recherche as part of the "Investissements d'avenir" program, reference ANR-19-P3IA-0001 (PRAIRIE 3IA Institute), Louis Vuitton ENS Chair on Artificial Intelligence, the European project MEMMO (Grant 780684) and the US National Science Foundation (grants 1932187 and 2026479).
\medskip
\newpage

\printbibliography
\appendix
\newpage
\section{Implementation details}

For each experiment, the data are normalized. For image measurement, the pixels are taken between 0~and~1 while for other type of measurements, the data are normalized to be centered with a standard deviation equal to~1. The penalty coefficient is always initialized to~1 and increased only once after 200 epochs. For each training, we report, $\tilde{\alpha}$, the value to which the parameter is increased. Finally, for every training, we use Adam with the default parameters: betas=(0.9, 0.999). The initial learning rate is $10^{-3}$ and we apply two decays with a factor 10 at epoch 200 and epoch 600. For each benchmark with RKN, we use the code\footnote{\url{https://github.com/LCAS/RKN} (GNU licence)} provided by the authors.

\section{Simple Pendulum}

\paragraph{Dataset.} Each trajectory is generated with an initial condition sampled uniformly with $\theta \in [-\pi, \pi]$ and $ \dot \theta \in [-1, 1]$. On each image, we add a centered Gaussian noise with a standard deviation of 0.2 (given pixels between 0 and 1).

\paragraph{Transition model.}

\begin{enumerate}[(a)]
        \item Fully connected: three hidden layers of 512 neurons with ReLU activations.
        \item Locally linear: K = 32, $\beta$ is a neural network with one hidden layer of 1024 neurons with a ReLU activation.
\end{enumerate}

\paragraph{Observation model.} 
\begin{itemize}\itemsep0.5pt
    \item One linear layer with $32 \times 3 \times 3$ neurons with ReLU activation.
    \item A 2D transpose convolution with 16 channels, a 5 by 5 kernel and a 3 by 3 stride. 
    \item A layer normalization \cite{ba2016layer} followed by a ReLU activation.
    \item A 2D transpose convolution with 12 channels, a 3 by 3 kernel and a 2 by 2 stride.
    \item A layer normalization followed by a ReLU activation.
    \item A 2D transpose convolution with 1 channel, a 2 by 2 kernel and a 1 by 1 stride. 
    \item A Sigmoid activation.
\end{itemize}

\paragraph{CNN-LSTM.}
\begin{itemize}\itemsep0.5pt
\item CNN: Two layers made of one convolution with 16 channels, a layer normalization, a ReLU activation and a 2 by 2 Maxpooling. And a final convolution layer with 16 channels.
\item LSTM with a hidden state of size of 1024.
\end{itemize}

\paragraph{Parameters.} We chose $\tilde{\alpha} = 10^3$ and train the SSM and the RNN for 1000 epochs with a batch size of~40.

\section{Quadruple Pendulum}
\paragraph{Dataset.} Each trajectory is generated with an initial condition sampled uniformly with $\theta_1, \theta_2 \in [-\pi, \pi]$. Each initial velocity is set to zero and we consider a damping of 0.1. On each image, we add a centered Gaussian noise with a standard deviation of 0.2 (given pixels between 0 and 1).

\paragraph{Transition model.}
\begin{enumerate}[(a)]
    \item Fully connected: three hidden layers of 1024 neurons with ReLU activations.
    \item Locally linear: K = 32, $\beta$ is a neural network with one hidden layer of 1024 neurons with a ReLU activation.
\end{enumerate}

\paragraph{Observation model.} 
\begin{itemize}\itemsep0.5pt
        \item One linear layer with $16 \times 3 \times 3$ neurons with ReLU activation.
        \item A 2D transpose convolution with 16 channels, a 5 by 5 kernel and a 3 by 3 stride. 
        \item A layer normalization followed by a ReLU activation.
        \item A 2D transpose convolution with 16 channels, a 3 by 3 kernel and a 2 by 2 stride.
        \item A layer normalization followed by a ReLU activation.
        \item A 2D transpose convolution with 12 channels, a 3 by 3 kernel and a 2 by 2 stride.
        \item A layer normalization followed by a ReLU activation.
        \item A 2D transpose convolution with 1 channel, a 2 by 2 kernel and a 1 by 1 stride. 
        \item A Sigmoid activation.
\end{itemize}

\paragraph{CNN-LSTM.}
\begin{itemize}\itemsep0.5pt
        \item CNN: Three layers made with one convolution with 16 channels, a layer normalization, a ReLU activation and 2 by 2 Maxpooling. And a final convolution layer with 16 channels.
        \item LSTM with a hidden state of size of 1024.
\end{itemize}

\paragraph{Parameters.} As for the simple pendulum, we chose $\tilde{\alpha} = 10^4$ and both the SSM  and the RNN are trained for 100 epochs with a batch size of 40.

\section{Moving MNIST}

\paragraph{Dataset.} The training set is generated by sampling MNIST digit and creating trajectories with random initial position and velocity. To do so, we use the implementation provided by \cite{hsieh2018learning}.

\paragraph{Transition model.} Locally linear with K = 32 and where $\beta$ is a neural network with one hidden layer of 1024 neurons with a ReLU activation.

\paragraph{Observation model.}
\begin{itemize}\itemsep0.5pt
        \item One linear layer with $512 \times 3 \times 3$ neurons with ReLU activation.
        \item A 2D transpose convolution with 512 channels, a 3 by 3 kernel and a 2 by 2 stride. 
        \item A layer normalization followed by a ReLU activation.
        \item A 2D transpose convolution with 256 channels, a 3 by 3 kernel and a 2 by 2 stride.
        \item A layer normalization followed by a ReLU activation.
        \item A 2D transpose convolution with 256 channels, a 3 by 3 kernel and a 2 by 2 stride.
        \item A layer normalization followed by a ReLU activation.
        \item A 2D transpose convolution with 1 channel, a 4 by 4 kernel and a 1 by 1 stride. 
        \item A Sigmoid activation.
\end{itemize}

\paragraph{CNN-LSTM.}
\begin{itemize}\itemsep0.5pt
    \item A convolutional layer with 64 channels
    \item A layer normalization followed by a ReLU activation and 2 by 2 Maxpooling.    
    \item A convolutional layer with 128 channels
    \item A layer normalization followed by a ReLU activation and 2 by 2 Maxpooling.     
    \item A convolutional layer with 512 channels
    \item A layer normalization followed by a ReLU activation and 2 by 2 Maxpooling.    
    \item A convolutional layer with 16 channels and a ReLu activation
    \item LSTM with a hidden state of size of 1024.
\end{itemize}

\paragraph{Parameters.} We chose $\tilde{\alpha} = 10^4$ and train the state space model for 250 epochs and then train, the RNN for 25 epochs. Both training are done with a batch size of 50.

\section{Lorenz system}

\paragraph{Dataset.} Each trajectory is generated with an initial condition sampled uniformly with $x_1, x_2, x_3 \in [-10, -10]$. Each trajectory is corrupted by a Gaussian noise with a standard deviation of $2.5$

\paragraph{Transition model.} Locally linear with K = 32 where $\beta$ is a neural network with one hidden layer of 1024 neurons with a ReLU activation.
    
\paragraph{Unscented Kalman filter.} We implement with Pytorch the UKF in the additive noise case \cite{wan2000unscented}. We initialize the covariance of $x_0$ at $0.1 I$ and chose a measurement noise variance of $0.5$ and a process noise variance of $10^{-6}$. The UKF is run on the normalized data, however, the metric is computed on the data without normalization.
    
\paragraph{RNN.} A LSTM with a hidden state of size of 1024.

\paragraph{Parameters.} We chose $\tilde{\alpha} = 10^2$ and train the SSM and the RNN for 1000 epochs with a batch size of~40.

\section{Training loss}

\begin{figure}[ht]
\begin{center}
\includegraphics[trim=10 8 10 20, clip, width=0.55\columnwidth]{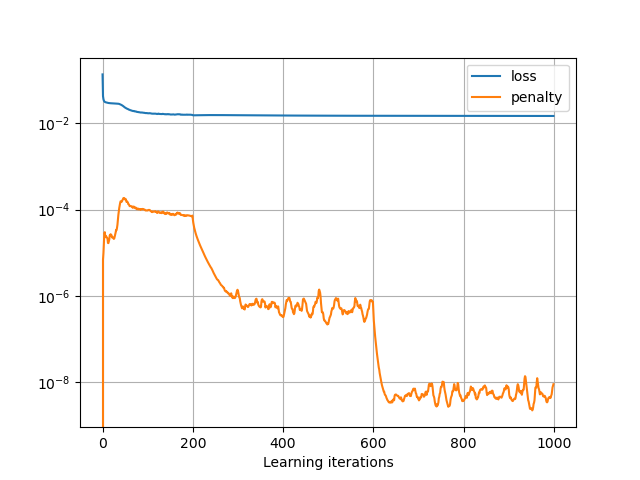}
\caption{Training loss. The blue curve corresponds to the fitting loss and the orange curve to the penalty term. }
\label{trainingloss}
\end{center}
\end{figure}

In Figure \ref{trainingloss}, we plot a typical evolution of the training loss on the simple pendulum (note that qualitatively similar training curves are seen for the other experiments). The total loss is the sum of both terms with the appropriate scaling of the penalty term. We can see that the two learning rate decays are especially useful to close the gaps. As expected, the loss does not converge to zero because of the measurement noise. The training of the SSM took 20 minutes on a GPU Quadro RTX 8000.

\section{Extra figures}

\begin{figure*}[ht]
\begin{center}
\subfigure[Ground truth]{\includegraphics[scale=0.16]{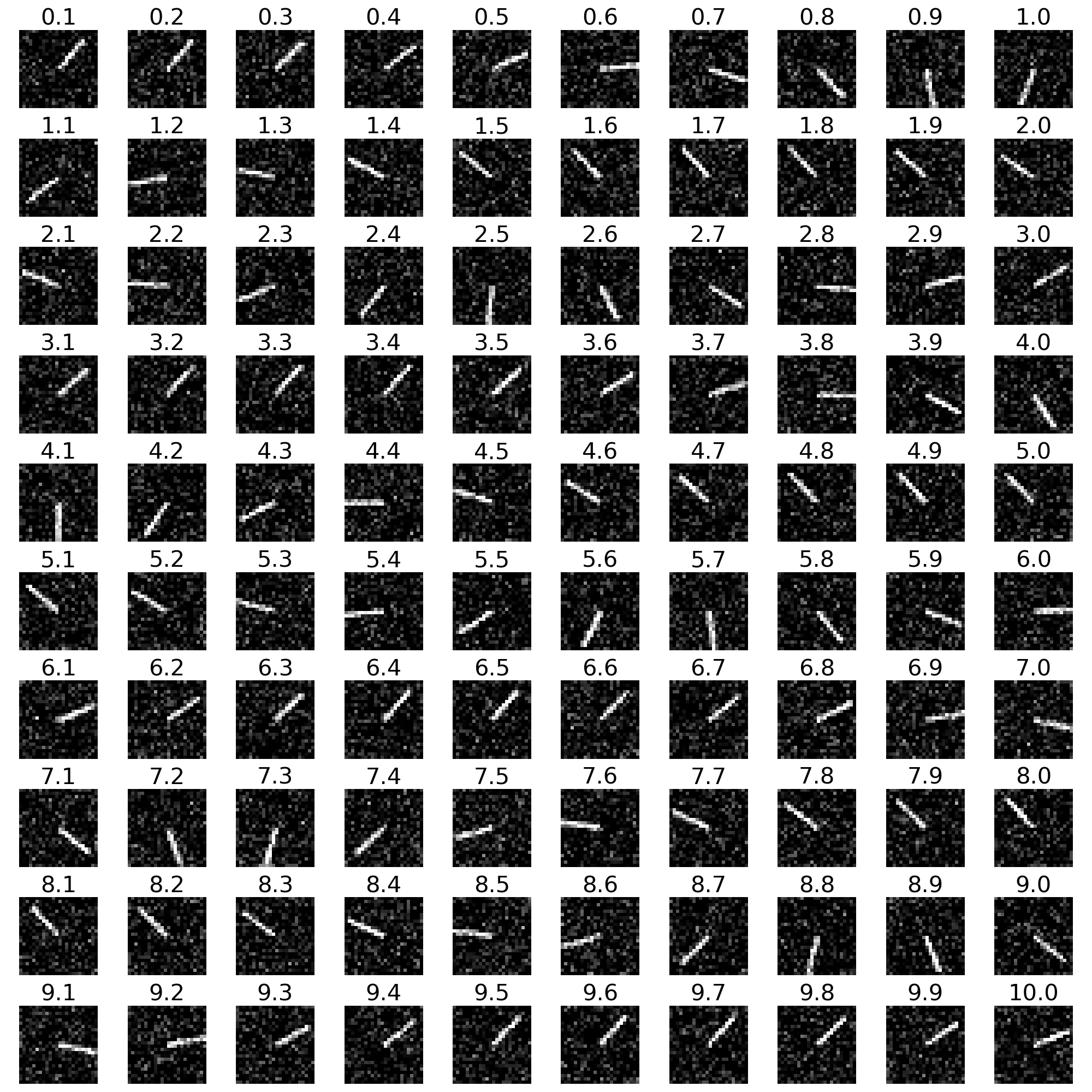}}\hspace{0.1cm}
\subfigure[Reconstruction of the filter and prediction]{\includegraphics[scale=0.16]{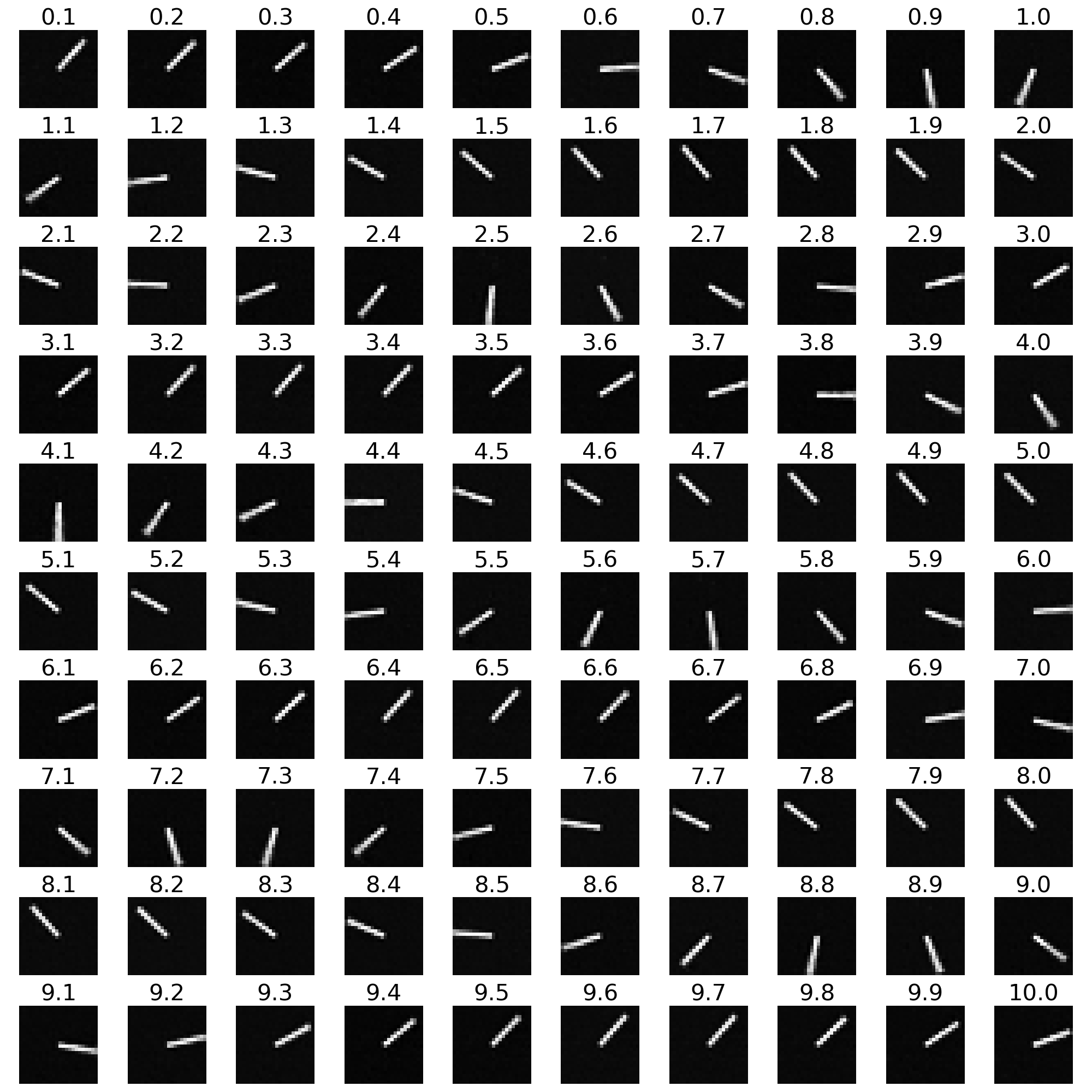}}
\caption{Prediction on the simple pendulum. The prediction starts at time 5 after the filtering process.}
\label{pendulum}
\end{center}
\end{figure*}

\begin{figure*}[ht]
\begin{center}
\subfigure[Ground truth]{\includegraphics[scale=0.16]{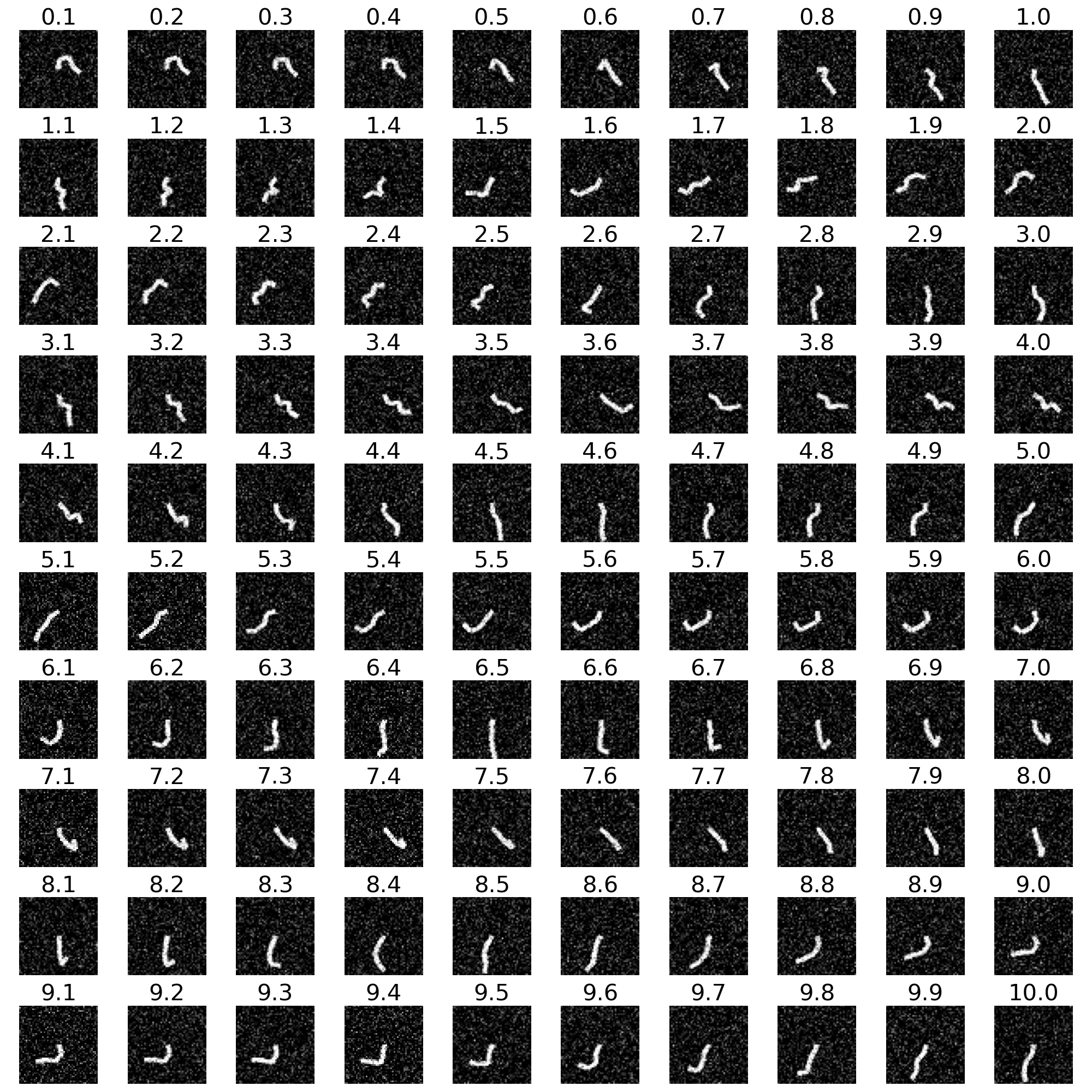}}\hspace{0.1cm}
\subfigure[Reconstruction of the filter and prediction]{\includegraphics[scale=0.16]{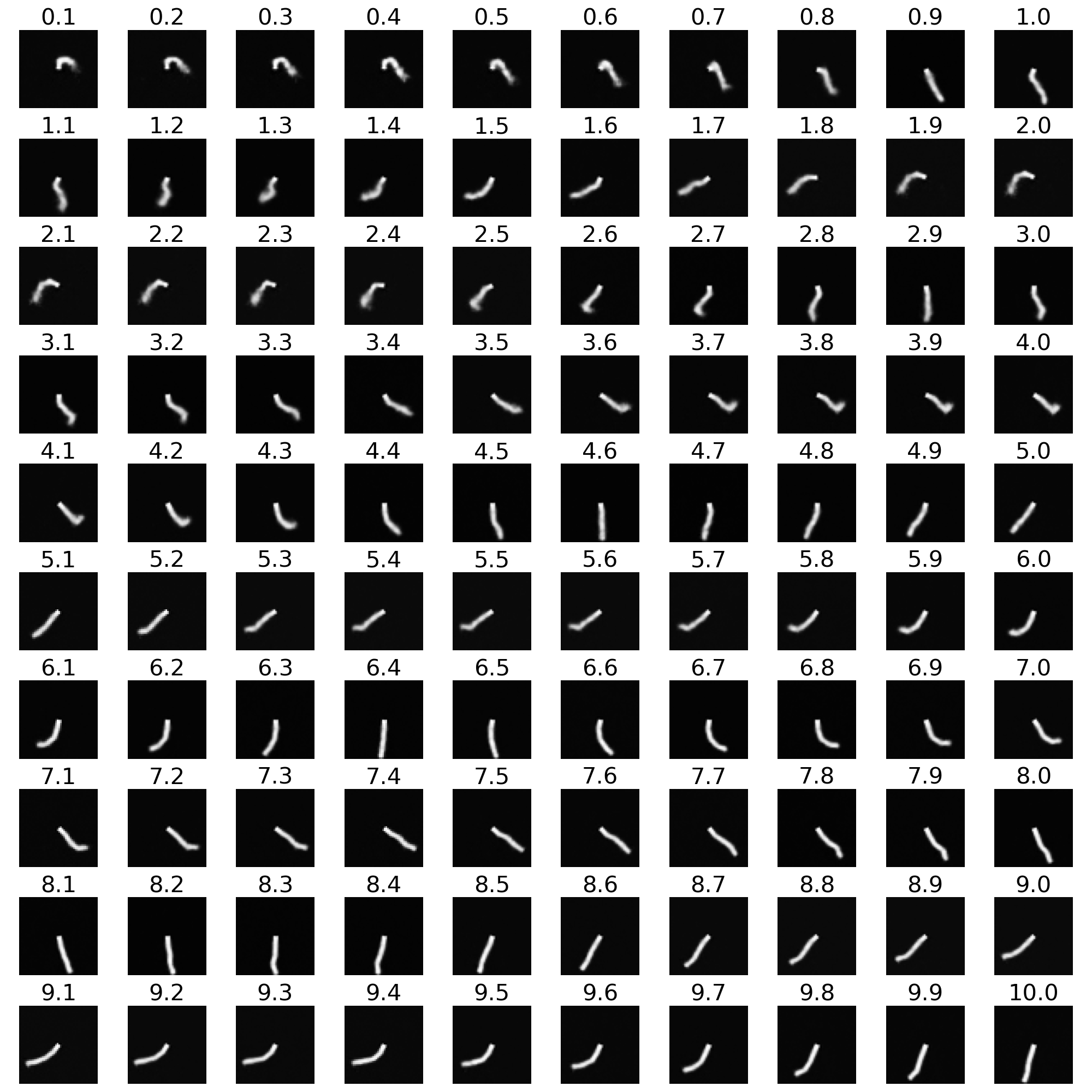}}
\caption{Prediction on the quadruple pendulum. The prediction starts at time 5 after the filtering process.}
\label{quad}
\end{center}
\end{figure*}

\begin{figure*}[ht]
\begin{center}
\subfigure{\includegraphics[trim=70 20 55 20, clip, width=\columnwidth]{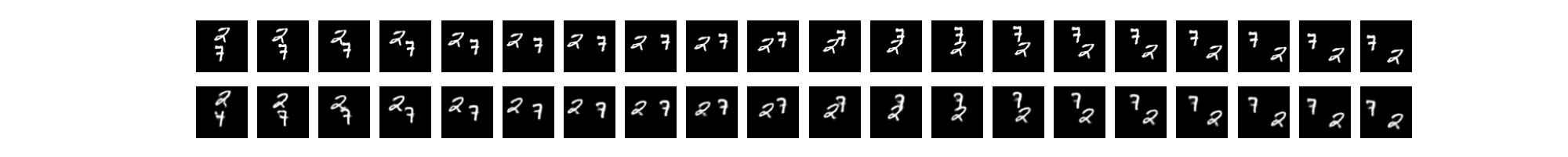}}
\vspace{0.3cm}
\subfigure{\includegraphics[trim=70 20 55 20, clip, width=\columnwidth]{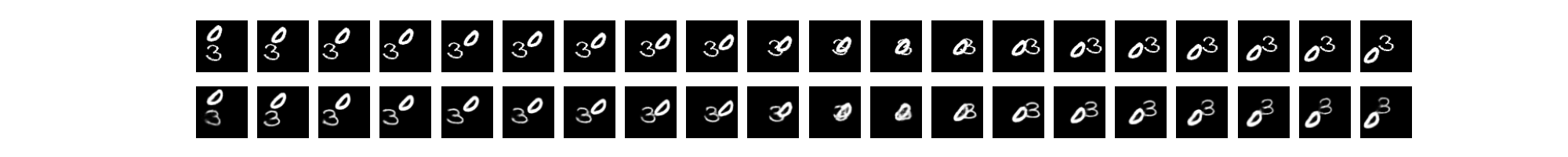}}
\vspace{0.3cm}
\subfigure{\includegraphics[trim=70 20 55 20, clip, width=\columnwidth]{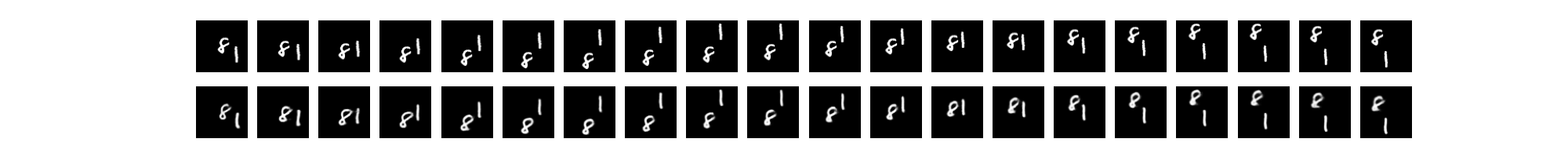}}
\vspace{0.3cm}
\subfigure{\includegraphics[trim=70 20 55 20, clip, width=\columnwidth]{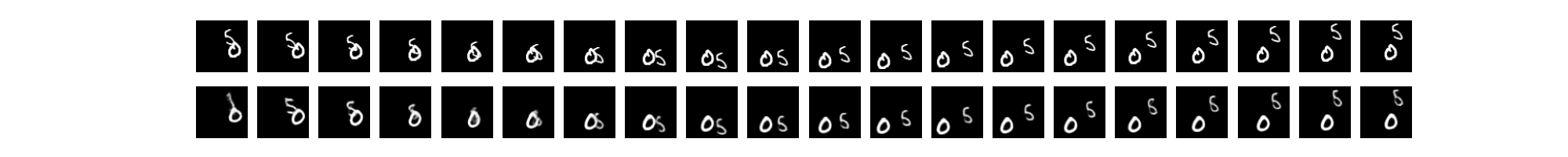}}
\vspace{0.3cm}
\subfigure{\includegraphics[trim=70 20 55 20, clip, width=\columnwidth]{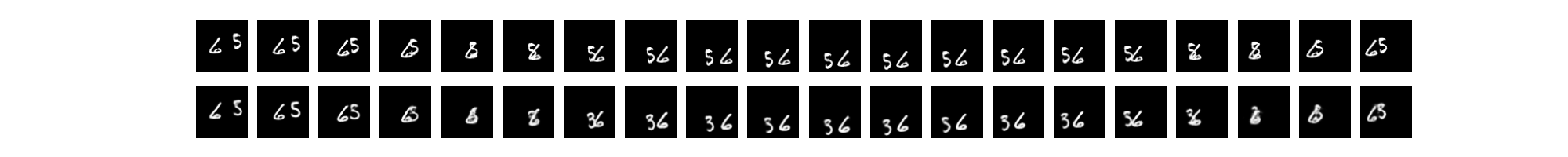}}
\vspace{0.3cm}
\subfigure{\includegraphics[trim=70 20 55 20, clip, width=\columnwidth]{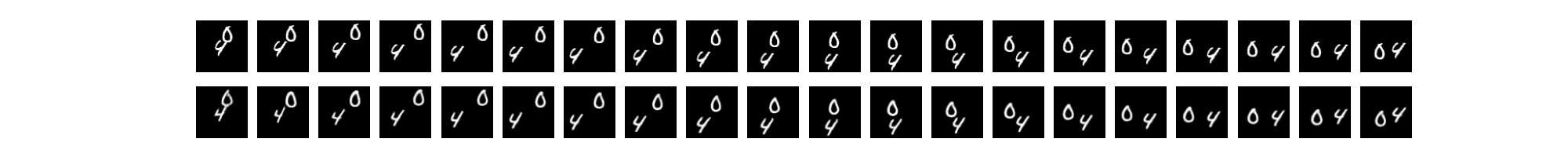}}
\vspace{0.3cm}
\subfigure{\includegraphics[trim=70 20 55 20, clip, width=\columnwidth]{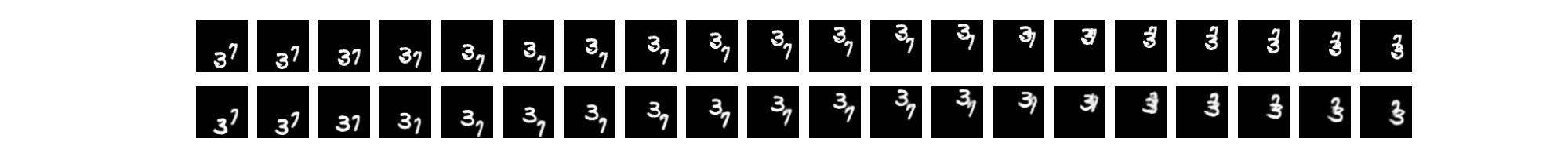}}
\vspace{0.3cm}
\subfigure{\includegraphics[trim=70 20 55 20, clip, width=\columnwidth]{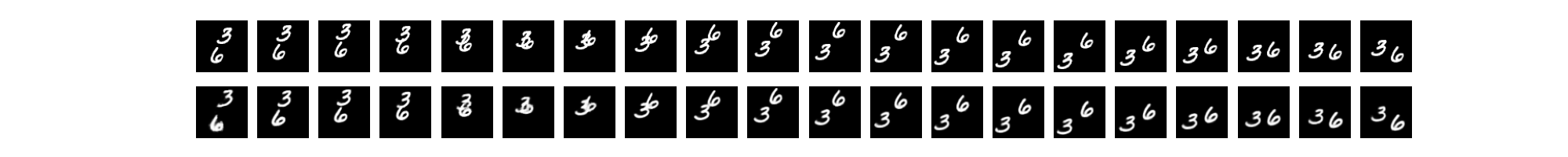}}
\vspace{0.3cm}
\subfigure{\includegraphics[trim=70 20 55 20, clip, width=\columnwidth]{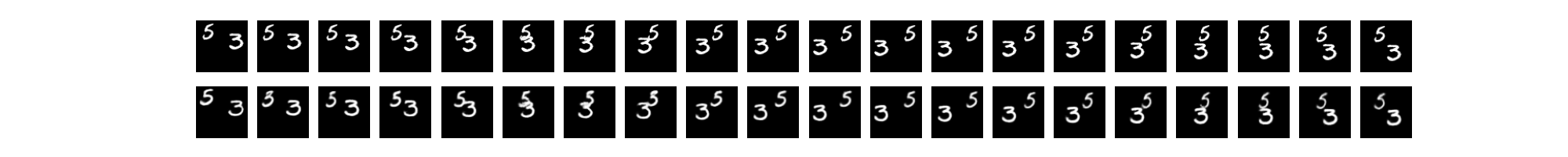}}
\vspace{0.3cm}
\subfigure{\includegraphics[trim=70 20 55 20, clip, width=\columnwidth]{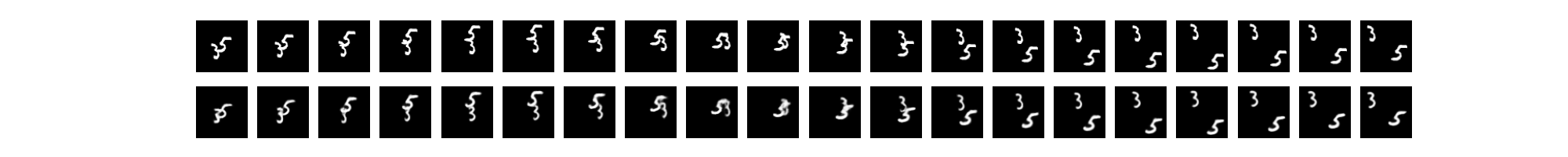}}
\caption{Various testing trajectories on Moving MNIST. Each first row corresponds to the ground truth and each second row to the belief of the filter and the prediction. The 10 first frames are filtering steps and the last 10 frames are prediction steps.}
\label{mnist}
\end{center}
\end{figure*}

\begin{figure*}[ht]
\begin{center}
\subfigure{\includegraphics[scale=0.42]{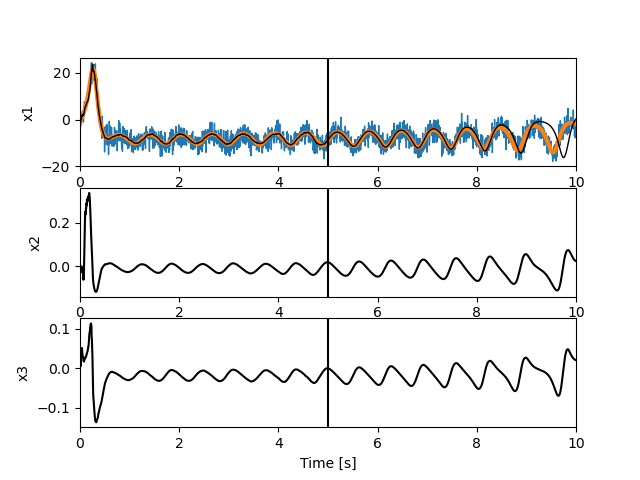}}
\subfigure{\includegraphics[scale=0.42]{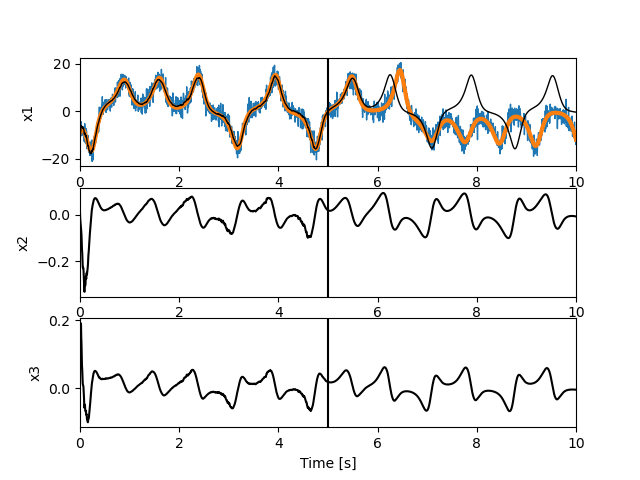}}
\subfigure{\includegraphics[scale=0.42]{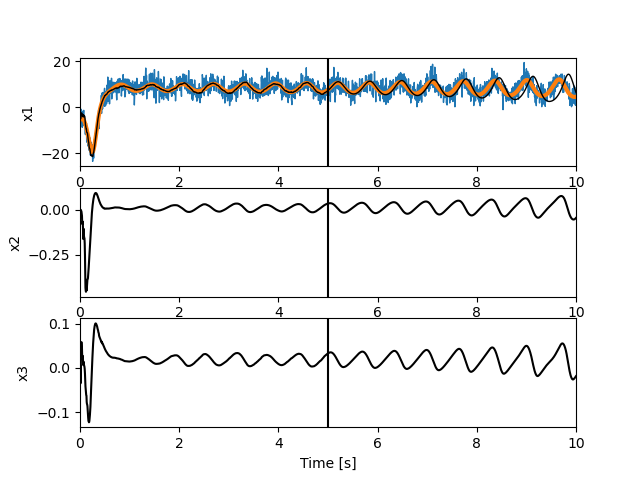}}
\subfigure{\includegraphics[scale=0.42]{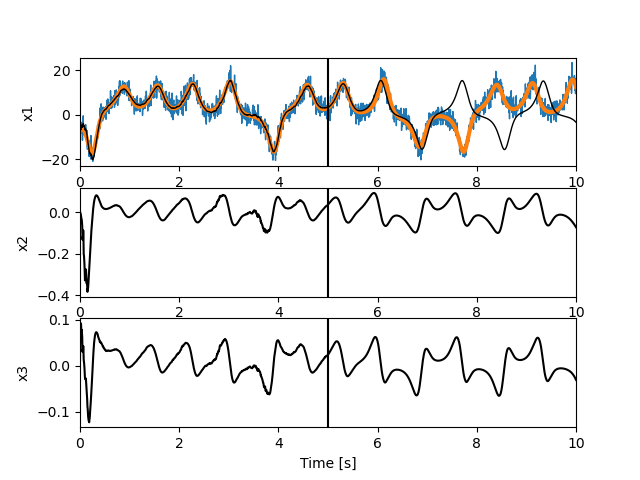}}
\caption{Various predictions on the Lorenz system. After 5 seconds of UKF filtering, the prediction starts. Each axis corresponds to one component of the state where the first one is also the measurement. The blue curve corresponds to the measurement, the orange curve to the measurement without noise and the black curve to the state.}
\label{lorenz_noise}
\end{center}
\end{figure*}

\begin{figure*}[ht]
\begin{center}
\subfigure{\includegraphics[scale=0.42]{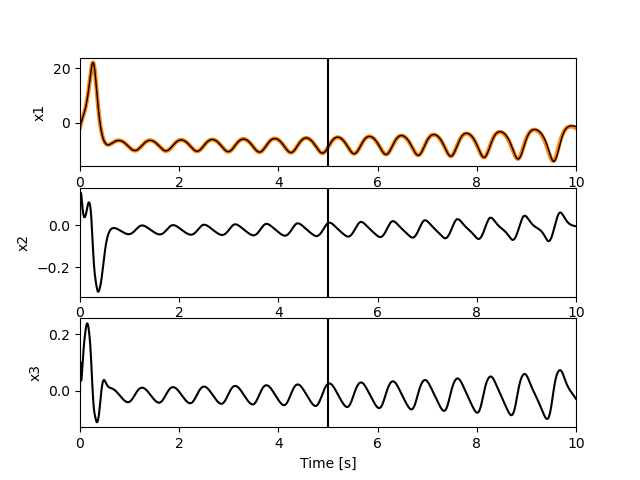}}
\subfigure{\includegraphics[scale=0.42]{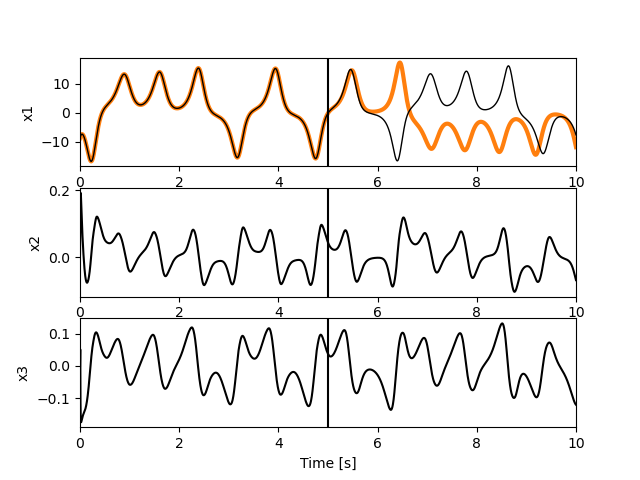}}
\subfigure{\includegraphics[scale=0.42]{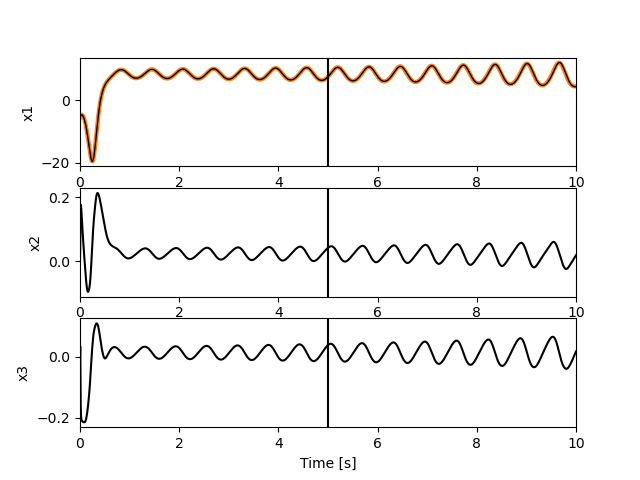}}
\subfigure{\includegraphics[scale=0.42]{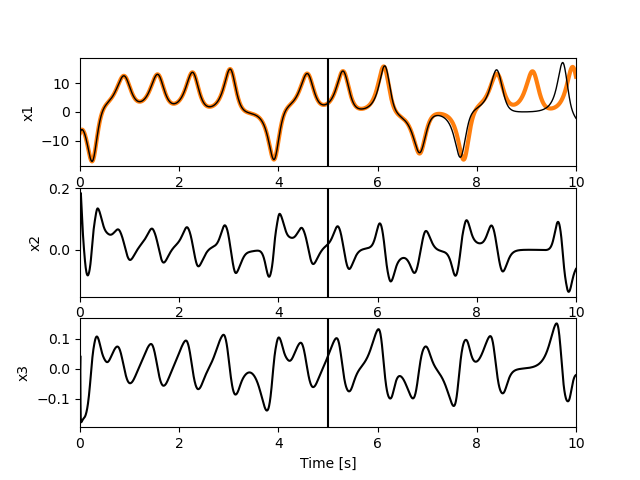}}
\caption{Various predictions on the Lorenz system without noise. After 5 seconds of UKF filtering, the prediction starts. Each axis corresponds to one component of the state where the first one is also the measurement. The orange curve corresponds to the measurement without noise and the black curve to the state.}
\label{lorenz_noiseless}
\end{center}
\end{figure*}

In Figure \ref{pendulum}, \ref{quad} and \ref{mnist}, we show complete predictions on the simple pendulum, the quadruple pendulum and Moving MNIST dataset. In Figure \ref{lorenz_noise}, we plot several testing trajectories on the Lorenz system. We can see that we can capture long term dynamics. Because of the chaotic nature of the system, once the prediction drifts, the error grows exponentially, especially at the change of orbits. However, it is important to notice that, even when the prediction diverges from the ground truth, our model remains stable and produces a realistic prediction even after 1000 steps (best exemplified by the reproduction of the strange attractor shown in the paper). Additionally, in Figure \ref{lorenz_noiseless}, we show predictions of a model trained without noise. The model can make very accurate prediction even after 1000 steps.

\end{document}